\documentclass[a4paper]{article}

\usepackage[utf8]{inputenc}
\usepackage{xcolor}
\usepackage{amsmath}
\usepackage{amsfonts}
\usepackage[tight]{subfigure}
\usepackage{fancyvrb}

\usepackage{graphicx}
\usepackage[sectionbib]{natbib}
\usepackage{authblk}

\newenvironment{keyword}
{\noindent\textbf{Keywords:} }
{\par}

\usepackage{hyperref}
\begin{document}

\title{Improving path-tracking performance of an articulated tractor-trailer system using a non-linear kinematic model}

\author[1]{Marina Murillo}
\author[1]{Guido Sanchez}
\author[1]{Nestor Deniz}
\author[1]{Lucas Genzelis}
\author[1]{Leonardo Giovanini}

\affil[1]{Research Institute for Signals, Systems and Computational Intelligence, sinc(i), FICH-UNL/CONICET, Ciudad Universitaria UNL, $4^\circ$ piso FICH, (S3000) Santa Fe, Argentina.}

\date{May 2022}

\maketitle

\begin{abstract}
This paper presents a novel non-linear mathematical model of an articulated tractor-trailer system that can be used, in combination with receding horizon techniques, to improve the performance of path tracking tasks of articulated systems. Due to its dual steering mechanisms, this type of vehicle can be very useful in precision agriculture, particularly for seeding, spraying and harvesting in small fields. The articulated tractor-trailer system model was embedded within a non-linear model predictive controller and the trailer position was monitored. When the kinematic of the trailer was considered, the deviation of trailer's position was reduced substantially alongside not only straight paths but also in headland turns. Using the proposed mathematical model, we were able to control the trailer's position itself rather than the tractor's position. The Robot Operating System (ROS) framework and Gazebo simulator were used to perform realistic simulations examples. 

\end{abstract}

\section{Introduction}
Precision agriculture (PA) is the art of merging high technology with agricultural machinery. The concept of PA is not new, however, in the last decades, its use among farmers has seen a rise due to improvements and low-cost development of electronics devices and high quality sensors, which allow the implementation of advanced control and signal processing algorithms

Tractors for agriculture purposes have been used along the 20th century. Indeed, after the second half of the 20th century they were continuously improved to be more efficient, productive and user-friendly. Farm machinery includes not only tractors but also transport vehicles, tillage and seeding machines, fertilizer applicators, and harvesters, among others. Due to mechanization and automation of these agricultural equipment, the intervention of human operators has been reduced. However, in most cases deviations from a desired trajectory are not corrected autonomously and the operator has to steer the vehicle in order to reduce the error. In order to relieve the operator of continuously making steering adjustments, several autonomous guidance systems for agricultural machinery have been developed 
\citep{baillie2018review,SUBRAMANIAN2006130,NAGASAKA2004223}.

One important automation problem that many applications have in common is the challenge of autonomous navigation of agricultural vehicles with towed implements. Generally, guidance systems control the trajectory of the vehicle so as to keep it as closer as possible to the desired path. However, when agricultural implements are used it would be more accurate to monitor its position rather than the tractor's because especially in curves and headland turns, the trailer tends to follow a different path leading to gaps and overlaps. Several works tackle the problem of controlling both the position of the tractor and the implement. For instance, \citet{pickett2016system} propose a system and method for steering an implement which enhances the potential tracking errors in the implement path on a sloped terrain. Both the vehicle and the implement have their own steering controller which steers both the vehicle and the implement steerable wheels in order to guide the implement towards the desired path. 
\citet{merx2017arrangement} present an arrangement that comprises a self-propelled vehicle with a towed implement. Here, the vehicle is capable of steering its own wheels and the implement can change its position in a lateral direction by means of an actuator coupled to the hitch point. Although in these works separate controllers for tractor and implement are used and a measure of the implement error is taken into account as an offset value, the main disadvantage of these solutions is that deviations from the nominal path caused by the tractor navigation, and vice versa, might not not be taken into account when navigating the trailer.  
\citet{kremmer2020system} propose a system and method for controlling an implement towed to an agricultural vehicle. Here, an actuator is mounted between the rear part of the chassis and the implement's hitch-point, thus allowing to move the whole implement in a parallelogram-wise manner in a lateral direction. As the controller proposed in this work is based on PID algorithm, it might be difficult to handle information regarding changes in road conditions and physical constraints of the system.

Agricultural vehicles with towed implements are not simple to control as they comprise highly non-linear dynamics and multiple inputs and outputs. In this regard, the use of modern control techniques such as model predictive control (MPC) for linear and non-linear systems (NMPC) have emerged \citep{rawlings2017model}.  
For instance, \citet{backman2012navigation} propose an NMPC method for a tractor and implement system. The main goal of their research was to control the lateral position of the towed implement and to keep it close to the adjacent driving line. The position of the implement was controlled by steering the tractor and by the use of a hydraulically controlled joint. 
\citet{kayacan2014learning} combine a fast centralized NMPC method based on ACADO code generation tool \citep{houska2011acado}, with nonlinear moving horizon estimation (NMHE) to obtain accurate trajectory tracking of an autonomous tractor-trailer system under unknown and variable soil conditions. 

\begin{figure}[h!]
\centering
\subfigure[]{
    \includegraphics[width=0.35\columnwidth]{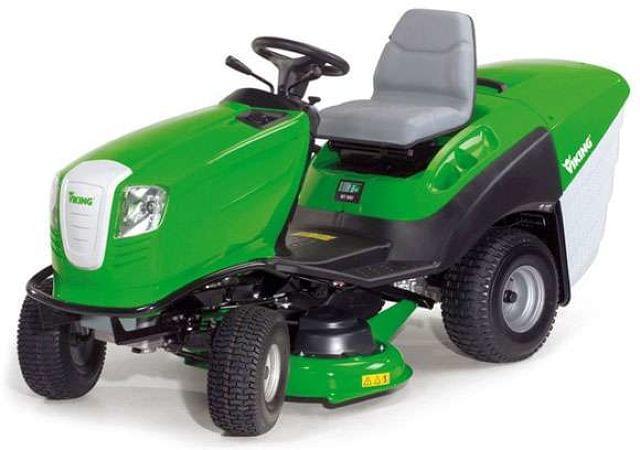}
    \label{fig:direccion_delantera}
    }
\subfigure[]{
    \includegraphics[width=0.35\columnwidth]{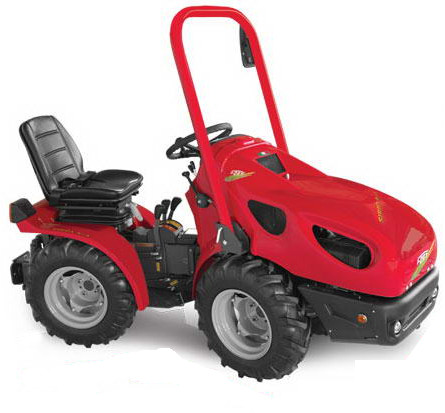}
    \label{fig:articulado}
    }    
\subfigure[]{
    \includegraphics[width=0.66\columnwidth]{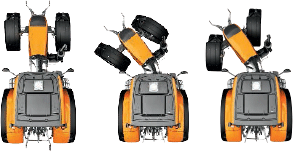}
    \label{fig:articulado_y_dobla}
    }    
\caption{Different turning mechanisms.}
\label{fig:tractores}
\end{figure}

On the other hand, tractors can change their orientation by means of two different kind of steering mechanisms. The most traditional one consists in steering the front wheels of the vehicle, as shown in Fig.~\ref{fig:direccion_delantera}\footnote{Source: \url{www.angliamowers.co.uk/viking-r5-mt-5097-z-garden-tractor.html}}. Another possibility is to provide the vehicle with a central articulated joint which is used for steering the vehicle instead of the traditional steering mechanism, as seen in Fig.~\ref{fig:articulado}\footnote{Source:  \url{www.fort-it.com/eng/agriculture-division/small-tractors/sirio}}. Although it is uncommon in the agricultural industry, both steering mechanisms can also be used within the same tractor, as it is depicted in Fig.~\ref{fig:articulado_y_dobla}\footnote{Source: \url{http://africa.valtra.com/en/articulated-tractors}}.

Since the performance of MPC-based controllers highly depends on the model describing the system behavior, a precise mathematical model is essential. The model embedded within the controller could be either kinematic or dynamic \citep{mondal2019comparison,tang2020improved}. While the first one deals with linear and angular speeds directly disregarding any inertia effects, the second one is concerned with forces and torques. The latter is usually more precise, however, it is mathematically more elaborate, thus, leading to controllers of greater computational complexity. Moreover, it might lead to numerical issues, affecting its implementation in different microcontrollers or single-board computers. In this regard, it has been shown that controllers based on kinematic models are accurate enough for vehicles operating at low accelerations \citep{werner2012kinematicVSdynamic, kong2015kinematic,tang2020improved}. 

Even though several articles dealing with the mathematical modeling of agricultural machinery can be found within the specialized literature, they mostly present simple models of tractors with front steering and they do not consider the kinematics of towed implements \citep{farmer2008kinematic-twobody,zhang2017robotics,nayl2013modeling}.  
There are other works which do consider vehicle-and-implement systems 
but these are limited to front-steering tractors \citep{kayacan2016centralized,yue2018robust}. In contrast, mathematical models of articulated vehicles have been published, but they do not incorporate the coupling of an implement nor front steering \citep{gustaf2012switching,gustaf2015effect}. 

As we plan drive vehicles at low speed, a kinematic model based controller would merely work well for us. To that end, in this article, we propose to study a kinematic tractor-trailer system model with both steering mechanisms: steering in the front wheels and a central articulated joint. It will be shown that, by restricting one steering mechanism or the other, the proposed model would suit any of the more limited cases. To the best of the authors' knowledge, neither the model presented in this article nor the technique used to derive it can be found in the specialized literature. This is the main contribution of this paper.

This work is organized as follows. In Section \ref{sec:modeling}, the derivation of a kinematic model of an articulated tractor-trailer system is carried out. A brief summary of the NMPC strategy is presented in Section \ref{sec:nmpc}. Section \ref{sec:pathfollowing} shows how the NMPC controller should be designed in order to guide the trailer's position alongside the desired trajectory. Simulation results using Gazebo\footnote{\url{http://gazebosim.org/}} simulator are depicted in Section \ref{sec:results}. The results obtained are thoroughly discussed in Section \ref{sec:discussion}. Finally, conclusions and future work are outlined in Section \ref{sec:concl}

\section{Articulated tractor-trailer system model}
\label{sec:modeling}
\begin{figure}[h!]
\centering
\includegraphics[scale=0.6]{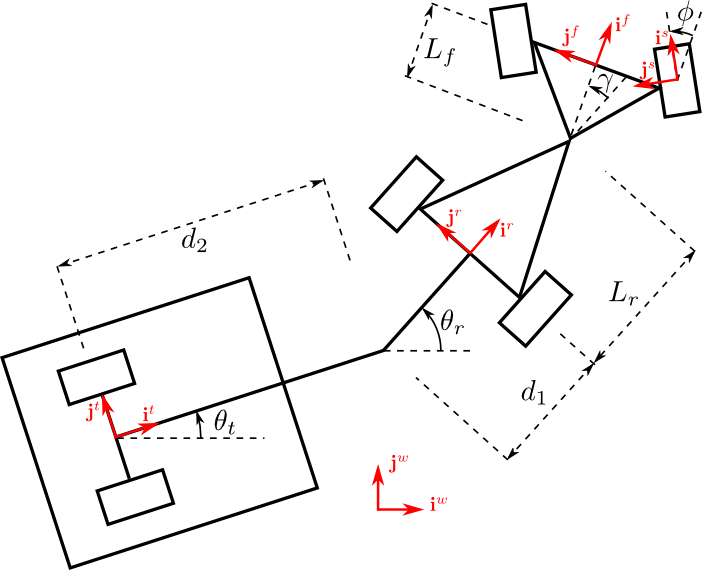}
\caption{Scheme of an articulated tractor-trailer system.}
\label{fig:modelo_articulado}
\end{figure}

A simple scheme of the proposed articulated tractor-trailer system is depicted in Fig.~\ref{fig:modelo_articulado}, where $L_r$ is the distance from the center of the rear axle of the tractor to the articulation joint, $L_f$ is the distance from this point to the center of the front axle, $d_1$ is the distance from the center of the rear axle to the trailer's hitch point, $d_2$ is the distance from this point to the center of the trailer's axle, $\theta_t$ is the trailer's yaw angle, $\theta_r$ is the yaw angle formed by the rear block of the tractor, $\gamma$ is the articulation angle and $\phi$ is the front steering angle.

In order to obtain a mathematical model of the system shown in Fig.~\ref{fig:modelo_articulado}, we have to consider five coordinate frames. In this figure unit vectors $\mathbf{i}$ and $\mathbf{j}$ corresponding to each reference system are also shown. The first coordinate frame is denoted with superscript $w$ and corresponds to the global reference frame, whose orientation is fixed. Frame $t$ matches the orientation of the trailer, i.e., vector $\mathbf{i}^t$ makes an angle $\theta_t$ with $\mathbf{i}^w$. Coordinate system $r$ matches the orientation of the rear part of the vehicle, and hence unit vector $\mathbf{i}^r$ makes an angle $\theta_r$ with $\mathbf{i}^w$. Frame $f$ has the same orientation as the front part of the vehicle, and therefore vector $\mathbf{i}^f$ makes an angle $\gamma$ with $\mathbf{i}^r$, that is, an angle $\theta_r + \gamma$ with $\mathbf{i}^w$. Finally, reference system $s$ matches the orientation of the front wheels, i.e., $\mathbf{i}^s$ makes an angle $\phi$ with $\mathbf{i}^f$, and thus an angle $\theta_r + \gamma + \phi$ with $\mathbf{i}^w$.

Let us proceed with the derivation of the mathematical model of the articulated tractor-trailer system by expressing the relationship between the location of the different parts of this system in terms of length constants and orientation angles previously defined. Let $[x_t, y_t]^T$, $[x_r, y_r]^T$ and $[x_f, y_f]^T$ be the position of the center of the trailer's axle, and the center of the tractor's rear axle and front axle, respectively, all expressed in the global frame $w$. Using the standard rotation matrix
\begin{equation}
    \label{eq:DCM}
    \text{R}(\theta) = 
     \begin{bmatrix} 
        \cos \theta & -\sin \theta \\ 
        \sin \theta & \cos \theta
    \end{bmatrix} ,
\end{equation}
the following geometric relationships can then be established:
\begin{subequations}
\begin{equation}
\begin{bmatrix} 
x_r \\ y_r
\end{bmatrix} = 
\begin{bmatrix} 
x_t \\ y_t
\end{bmatrix} +
\text{R}(\theta_t) \begin{bmatrix} 
d_2 \\ 0
\end{bmatrix} +
\text{R}(\theta_r) \begin{bmatrix} 
d_1 \\ 0
\end{bmatrix},
\label{eq:relac_geom1}
\end{equation}
\begin{equation}
\begin{bmatrix} 
x_f \\ y_f
\end{bmatrix} = 
\begin{bmatrix} 
x_r \\ y_r
\end{bmatrix} +
\text{R}(\theta_r) \begin{bmatrix} 
L_r \\ 0
\end{bmatrix} +
\text{R}(\theta_r+\gamma) \begin{bmatrix} 
L_f \\ 0
\end{bmatrix}.
\label{eq:relac_geom2}
\end{equation}
\end{subequations} 
The time-derivatives of Eqs.~(\ref{eq:relac_geom1}) and (\ref{eq:relac_geom2}) can be expressed as
\begin{equation}
\left\{ \begin{array}{l}
\dot x_r =  \dot x_t - d_2 \dot \theta_t \sin \theta_t - d_1 \dot \theta_r \sin \theta_r \\
\dot y_r =  \dot y_t + d_2 \dot \theta_t \cos \theta_t + d_1 \dot \theta_r \cos \theta_r \\
\dot x_f =  \dot x_r - L_r \dot \theta_r \sin \theta_r - L_f (\dot \theta_r + \dot \gamma) \sin (\theta_r +\gamma) \\
\dot y_f =  \dot y_r + L_r \dot \theta_r \cos \theta_r + L_f (\dot \theta_r + \dot \gamma) \cos (\theta_r +\gamma) \\
\end{array} \right. .
\label{eq:relac_geom_der}
\end{equation}

Assuming lateral slip cannot take place, each wheel is restricted to move in the longitudinal direction. However, if this constraint is imposed on each wheel individually, the vehicle would only be allowed to move in a straight line, i.e., with $\theta_t = \theta_r$ and $\gamma = \phi = 0$. Consequently, the model is further simplified treating the system as if each axle had a single wheel located on its center. This simplification is commonly referred to as ``bicycle model" and it is commonplace in the modeling of ground vehicles \citep{zhang2017robotics, lavalle2006planning, corke2001steering, siew2009tractorimplementtrailer}. Using this simplification, we allow each block of the system (trailer, rear part and front part) to move only in the direction orthogonal to its axle. These non-holonomic constraints can be expressed as
\begin{equation}
\begin{array}{lll}
\begin{bmatrix} \dot x_t \\ \dot y_t \end{bmatrix} = \text{R}(\theta_t) \begin{bmatrix} v_t \\ 0 \end{bmatrix}, \hspace{0.1cm} &
\begin{bmatrix} \dot x_r \\ \dot y_r \end{bmatrix} = \text{R}(\theta_r) \begin{bmatrix} v_r \\ 0 \end{bmatrix} \hspace{0.1cm} \text{and} \hspace{0.1cm} & 
\begin{bmatrix} \dot x_f \\ \dot y_f \end{bmatrix} = \text{R}(\theta_r + \gamma + \phi) \begin{bmatrix} v_f \\ 0 \end{bmatrix},
\end{array}
\end{equation}
where $v_t$, $v_r$ and $v_f$ are the speeds of the center of the trailer axle, rear axle and front axle, respectively. It is worth noting that the angle $\theta_r + \gamma + \phi$ was used instead of $\theta_r + \gamma$ so as to take into account the tractor's front steering. Working with these expressions and putting them all together yields the following equalities:
\begin{equation}
\left\{ \begin{array}{l}
\dot x_t = v_t \cos \theta_t \\
\dot y_t = v_t \sin \theta_t \\
\dot x_r = v_r \cos \theta_r \\
\dot y_r = v_r \sin \theta_r \\
\dot x_f = v_f \cos (\theta_r + \gamma + \phi) \\
\dot y_f = v_f \sin (\theta_r + \gamma + \phi)
\end{array} \right. .
\label{eq:rest_no_holonomicas}
\end{equation}
Replacing these relationships in Eqs.~(\ref{eq:relac_geom_der}) results in:
\begin{equation}
\hspace{-1em}
\left\{ \begin{array}{l}
\begin{array}{lr}
\hspace{-1em} v_r  \cos \theta_r  =  v_t \cos \theta_t - d_2 \dot \theta_t \sin  \theta_t - d_1 \dot \theta_r \sin \theta_r & \hspace{1em} (a) \hspace{-5em}\\
\hspace{-1em} v_r  \sin \theta_r   = v_t  \sin \theta_t + d_2 \dot \theta_t \cos \theta_t + d_1 \dot \theta_r \cos \theta_r &\hspace{1em} (b) \hspace{-5em}\\
\end{array} \\
\begin{array}{llr}
\hspace{-1em} v_f  \cos (\theta_r + \gamma + \phi) & = v_r \cos \theta_r  - L_r \dot \theta_r \sin \theta_r \\ 
\hspace{-1em} & - L_f (\dot \theta_r + \dot \gamma) \sin (\theta_r +\gamma)  & (c) \\
\hspace{-1em} v_f  \sin (\theta_r + \gamma + \phi) & = v_r \sin \theta_r + L_r \dot \theta_r \cos \theta_r \\
\hspace{-1em} & + L_f (\dot \theta_r + \dot \gamma) \cos (\theta_r +\gamma) & \hspace{1em} (d) \\
\end{array} 
\end{array} \right. .
\label{eq:restric_reempl}
\end{equation}
Multiplying Eq.~(\ref{eq:restric_reempl}c) by $-\sin \theta_r$ and Eq.~(\ref{eq:restric_reempl}d) by $\cos \theta_r$, and then adding the resulting expressions together, it can be shown that
\begin{equation}
\dot \theta_r = \frac{v_f \sin(\gamma + \phi) - \dot \gamma L_f \cos \gamma}{ L_r + L_f \cos \gamma}.
\label{eq:theta_r}
\end{equation}
As it can be easily seen, this expression would cause problems if 
\begin{equation}
L_r + L_f \cos \gamma = 0,    
\end{equation}
However, due to mechanical limitations of articulated-tractors, $\gamma$ is limited to $-\frac{\pi}{2} < \gamma < \frac{\pi}{2}$, therefore $\cos \gamma \geq 0 $ and this difficulty will not arise. 
Similarly, multiplying Eq.~(\ref{eq:restric_reempl}a) by $-\sin \theta_t$ and adding it to Eq.~(\ref{eq:restric_reempl}b) multiplied by $\cos \theta_t$ it yields
\begin{equation}
\dot \theta_t = \frac{v_r}{d_2} \sin(\theta_r - \theta_t) - \frac{d_1}{d_2} \dot \theta_r \cos(\theta_r - \theta_t) .
\label{eq:theta_t}
\end{equation}

Let us now proceed to define the control inputs and state variables for the system under study. Based on Eqs.~(\ref{eq:theta_r}) and (\ref{eq:theta_t}), it seems natural to consider angles $\theta_r$ and $\theta_t$ as state variables. Additionally, since Eq.~(\ref{eq:theta_r}) involves the time-derivative of $\gamma$, it is convenient to include this angle as another state variable. Setting the angular velocity of the articulation joint $\omega_1$ as a control input, it results in
\begin{equation}
\dot \gamma = \omega_1.
\label{eq:gamma}
\end{equation}
On the other hand, the time-derivative of the forward steering angle $\phi$ is not involved in any of the previous expressions. Hence, this angle could be considered either as a state variable or a control input. The latter allows for constraints on the rate of change of this angle to be easily incorporated into the control problem, leading to a smoother behavior of the system. Therefore, this second alternative has been chosen in this work. Defining the rate of change of $\phi$, $\omega_2$, as another control then
\begin{equation}
\dot \phi = \omega_2.
\label{eq:phi}
\end{equation}

\noindent In order to fully specify the system, the position of any of its blocks needs to be known. Given that it is of interest to control the position of the trailer, $x_t$ and $y_t$ are selected as state variables. Using Eqs.~(\ref{eq:restric_reempl}a), (\ref{eq:restric_reempl}b) and (\ref{eq:rest_no_holonomicas}), it can be easily shown that
\begin{equation}
\left\{ \begin{array}{l}
\dot x_t = v_r  \cos \theta_r  + d_2 \dot \theta_t \sin  \theta_t + d_1 \dot \theta_r \sin \theta_r \\
\dot y_t = v_r  \sin \theta_r - d_2 \dot \theta_t \cos \theta_t - d_1 \dot \theta_r \cos \theta_r \\
\end{array} \\
\right. .
\label{eq:pos_t} 
\end{equation}

\noindent Finally, the speed of either the rear or the front block of the tractor, i.e. $v_r$ or $v_f$, must be defined as the last control input. 
In this work $v_f$ has been chosen, so as to pose a more challenging control problem, since in this way the chain of mechanisms acting between the trailer and the directly-actuated block of the tractor is longer. 
The complete kinematic model of the articulated tractor-trailer system can be obtained by grouping together Eqs.\ (\ref{eq:theta_r}) - (\ref{eq:pos_t}), yielding
\begin{equation}
\begin{bmatrix}
    \dot x_t \\
    \dot y_t \\
    \dot \theta_r \\
    \dot \theta_t \\
    \dot \gamma \\
    \dot \phi 
\end{bmatrix} = 
\begin{bmatrix}
    v_r  \cos \theta_r + d_1 \dot \theta_r \sin \theta_r  + d_2 \dot \theta_t \sin  \theta_t \\
    v_r  \sin \theta_r - d_1 \dot \theta_r \cos \theta_r - d_2 \dot \theta_t \cos \theta_t \\
    \frac{v_f \sin(\gamma + \phi) - \omega_1 L_f \cos \gamma}{ L_r + L_f \cos \gamma} \\
    \frac{v_r}{d_2} \sin(\theta_r - \theta_t) - \frac{d_1}{d_2} \dot \theta_r \cos(\theta_r - \theta_t) \\
    \omega_1 \\
    \omega_2
\end{bmatrix}
\label{eq:sist_full_cin} 
\end{equation}
where $v_r$ can be obtained as  
\begin{equation}
v_r = v_f \cos (\gamma + \phi) + L_f (\dot \theta_r + \dot \gamma) \sin \gamma,    
\end{equation}
and $\dot{\theta}_r$ and $\dot{\theta}_t$ are defined in Eqs.~(\ref{eq:theta_r}) and (\ref{eq:theta_t}). 

\noindent Defining 
\begin{equation}
\mathbf{x} = [x_t, y_t, \theta_r, \theta_t, \gamma, \phi]^T \hspace{0.3cm} \text{and} \hspace{0.3cm} \mathbf{u} = [v_f, \omega_1, \omega_2]^T 
\label{eq:state_input_vectors}
\end{equation}
as our state and control input vectors, respectively, Eq.~(\ref{eq:sist_full_cin}) can be written in a compact vector-matrix form as
\begin{equation}
\mathbf{\dot x} = F(\mathbf{x},\mathbf{u}),
\label{eq:sist_vect} 
\end{equation}
where $F(\mathbf{x},\mathbf{u})$ is the vector function given by the right hand side (RHS) of Eq.~(\ref{eq:sist_full_cin}). It is worth mentioning that we decided to choose the state vector $\mathbf{x}$ as defined in Eq.~(\ref{eq:state_input_vectors}) because we need to know the position and orientation of the trailer. In this regard, $x_t$ and $y_t$ define the trailer’s \emph{xy}-position and $\theta_t$ is the trailer’s yaw angle. The other three angles ($\theta_r$, $\gamma$, and $\phi$) are directly related to the trailer’s position and orientation equations. It is interesting to note that the mathematical model we have obtained can be regarded as a generalization of many other models found in the specialized literature. For example, if the front direction is fixed ($\phi \equiv \omega_2 \equiv 0$) and the trailer is neglected, ignoring $\theta_t$ and replacing the equations for $\dot x_t$ and $\dot y_t$ with the corresponding equations for $\dot x_r$ and $\dot y_r$, the resulting system matches the one obtained by \citet{gustaf2015effect}. Moreover, if it is assumed that the hitch point of the trailer is located directly on the rear axle of the tractor ($d_1 = 0$) and the articulation joint is removed (setting $\gamma \equiv \omega_1 \equiv 0$), the model obtained matches the one presented by \citet{lavalle2006planning}. 

\section{Non Linear Model Predictive Control}
\label{sec:nmpc}
In order to show the advantages of using the mathematical model of the articulated tractor-trailer system described by Eq.~(\ref{eq:sist_full_cin}), we propose to use a model based control technique such as NMPC due to its high capabilities to deal with non-linear models and constraints. This technique is not new, however, as it will be shown in Section~\ref{sec:results}, by using our articulated tractor-trailer system model within a NMPC controller it is possible to address the problem of trailer’s path tracking in a precise way. Another advantage of using NMPC technique is that perturbations affecting the system can be added in the minimization stage, thus, the performance of the controller can be improved as the resulting control inputs take into account this new information. It should be pointed out that other techniques do not allow to do this in such an efficient and easy way as receding horizon techniques do.

The main purpose of NMPC is to predict the future states of the system solving an explicit inverse problem that allows the incorporation, at the design stage, of different types of constraints to obtain the best feasible solution. The inverse problem to be solved is the minimization of a cost function that quantifies the performance of the system. This constrained minimization process is done over a fixed-time horizon window of a length $N$. At the next sampling instant, new information is included and old one is discarded by shifting the window one step in time and the constrained minimization process is restarted at the next sampling instant \citep{rawlings2017model}. Generally, NMPC is implemented in discrete-time, hence the general form of the problem to be solved is
\begin{equation}
\begin{array}{c}
\underset{\mathbf{U}_{k|k}}{\operatorname{min}} \hspace{5pt} \mathcal{J}(k) \\[0.5cm]
\begin{array}{ll}
\text{st.} &  
    \left\{
    \begin{array}{l}
        \mathbf{x}_{k+i+1|k} = f(\mathbf{x}_{k+i|k},\mathbf{u}_{k+i|k}), \hspace{0.3cm} i \in [0,1,\cdots,N-1] \\
        \mathbf{x}_{k|k} = \mathbf{x}(k), \\
        \mathbf{u}_{k+i|k} \in \mathcal{U}, \hspace{0.3cm} \mathbf{x}_{k+i|k} \in \mathcal{X},
    \end{array}
    \right.
\end{array}
\end{array}
\label{eq:obj_fun_mpc}
\end{equation}
where $\mathcal{J}(k)$ denotes the cost function to be minimized, $\mathbf{x}_{k+i|k} \in \mathcal{X} \subseteq \Re ^{n_x}$ is the state vector, $\mathbf{u}_{k+i|k} \in \mathcal{U} \subseteq \Re ^{n_u}$ is the control input vector, $N$ is the control window length, 
$\mathcal{X}$ and $\mathcal{U}$ are the state and input constraint sets, respectively, $\mathbf{U}_{k|k}=\left[\mathbf{u}_{k|k}, \, \cdots, \, \mathbf{u}_{k+N-1|k}\right]^T$ is the control input sequence and $f(\cdot)$ is a vector function that describes the dynamics of the system. It is worth noting that subscript ${k+i|k}$ refers to the information computed at time $k+i$ using the information available at time $k$. The solution of the problem defined in Eq.~(\ref{eq:obj_fun_mpc}) is an optimal control input sequence $\mathbf{U}^*_{k|k}=\left[\mathbf{u}^*_{k|k}, \, \cdots, \, \mathbf{u}^*_{k+N-1|k}\right]^T$, but only the first control input of this sequence is applied to the system, i.e. $\mathbf{u}_{k}=\mathbf{u}^*_{k|k}$. Then, the horizon is shifted forward to the next sampling instant in a receding horizon fashion, discarding old information and including new one, thus compensating for unmeasured disturbances and/or unmodeled dynamics. As it can be seen, the cost function plays a key role in obtaining the optimal control sequence and it should be carefully designed in order to fulfill the goals of the system. 

Another benefit of using NMPC technique is that obstacles can indeed be considered within the controller. To that end, any obstacle can be modeled by a polytope\footnote{Note that the space occupied by the obstacle can also be described, roughly, by an ellipse to reduce the number of used constraints.}, which can be implemented through a set of linear constraints. Thus, adding an obstacle to the constrained minimization problem is just as simple as including a constraint of the form $g(x_t,y_t,x_{o},y_{o}) - \sigma \leq 0$, where $g$ and $\sigma$ describe the linear polytopic constraints, and $x_o$ and $y_o$ denote the \emph{xy}-coordinates of the obstacle. Since the obstacle is added as a constraint in Eq.~(\ref{eq:obj_fun_mpc}), its detection and avoidance is straightforward, because the solution of the optimization problem already takes into account the presence of this obstacle.

\section{Path-following with the articulated tractor-trailer system}
\label{sec:pathfollowing}
The goal of this section is to design a NMPC based controller for the articulated tractor-trailer system that allows to control the $xy$-position of the trailer along a predefined path. In order to use the NMPC technique we need a discrete-time model of the system, hence, we must discretize Eq.~(\ref{eq:sist_full_cin}). There are several non-linear discretization methods that can be used such as shooting method, Runge-Kutta method (among which the popular fourth-order explicit method can be found) and collocation method. The latter involves finding, for each discretization period, polynomials of a certain order that satisfy the system's differential equations in a specific set of points \citep{diehl2006fastmultipleshooting,milne1972handbook}, which can be obtained, for instance, from the Gauss-Legendre quadrature. In this work, collocation method will be used as it provides great accuracy at a relatively low computational cost \citep{sanchez2017discmethods_review}. In this way, Eq.~(\ref{eq:sist_vect}) can be transformed into its equivalent discrete-time as
\begin{equation}
\mathbf{x}_{k+1} = \hat F (\mathbf{x}_{k}, \mathbf{u}_{k}),
\label{eq:sist_vect_disc}
\end{equation}
where $\mathbf{x}_{k} = [x_{t_k}, y_{t_k}, \theta_{r_k}, \theta_{t_k}, \gamma_k, \phi_k]^T$ is the discrete-time state vector, $\mathbf{u}_k = [v_{f_k}, \omega_{1_k}, \omega_{2_k}]^T$ is the discrete-time control input vector and $\hat F(\mathbf{x}_{k},\mathbf{u}_{k})$ approximates the RHS of Eq.~(\ref{eq:sist_full_cin}) in discrete-time. 

A natural reference input for the controller would be the trajectory $\mathbf{r}_{\mathbf{x}_{\{x_t,y_t\}}}$ that should be followed by the trailer, where ${\mathbf{x}_{\{x_t,y_t\}}}$ means that from the state vector $\mathbf{x}$ only setpoints for states $x_t$ and $y_t$ are considered. Then, using these points as the desired \emph{xy}-position of the trailer, we propose to solve problem defined in Eq.~(\ref{eq:obj_fun_mpc}) with the following cost function:
\begin{equation}
\begin{array}{rl}
    \mathcal{J}(k) = & \sum\limits_{j=0}^{N-1} \Vert{\mathbf{x}_{{{\{x_t,y_t\}}}_{k+j|k}} - \mathbf{r}_{\mathbf{x}_{{\{x_t,y_t\}}_{k+j|k}}}}\Vert^2_Q + \Vert{\mathbf{u}_{k+j|k}}\Vert^2_R  \\ 
  & + \Vert{\mathbf{x}_{{\{x_t,y_t\}}_{k+N|k}} - \mathbf{r}_{\mathbf{x}_{{\{x_t,y_t\}}_{k+N|k}}}}\Vert^2_P
\end{array}
\label{eq:cost_etap_y_term_modif}
\end{equation}
where $\mathbf{x}_{{\{x_t,y_t\}}_{k+j|k}}$ denotes the discrete-time \emph{xy}-position of the trailer, $\mathbf{u}_{k+j|k}$ is the discrete-time control input vector of the articulated tractor-trailer system, $Q$, $P$ and $R$ are positive definite cost matrix and $N$ is the prediction horizon length. The last term in Eq.~(\ref{eq:cost_etap_y_term_modif}) is known as terminal cost as it summarizes the information between samples $N$ and $\infty$, which was not taken into account in the minimization problem because, in fact, we are solving a finite optimization problem rather than an infinite one. Moreover, if matrix $P$ is set accordingly, the terminal cost can also be used to guarantee the stability of the solutions.

\section{Simulation results}
\label{sec:results}
The simulation examples presented in this section were run within an Intel\textsuperscript{\textregistered} Core\textsuperscript{\texttrademark} i7-8700 CPU @ 3.20GHz with 16 GB RAM. The code was written using \textit{Python} and a symbolic framework for algorithmic differentiation and optimization named CasADi \citep{Andersson2018}, in conjunction with the toolbox ``Nonlinear Model Predictive Control Tools for CasADi" \citep{mpc-tools-casadi} and the HSL Mathematical Software Library \citep{hsl}. 

\begin{figure}[!h]
\centering
\subfigure{
    \includegraphics[width=0.5\columnwidth]{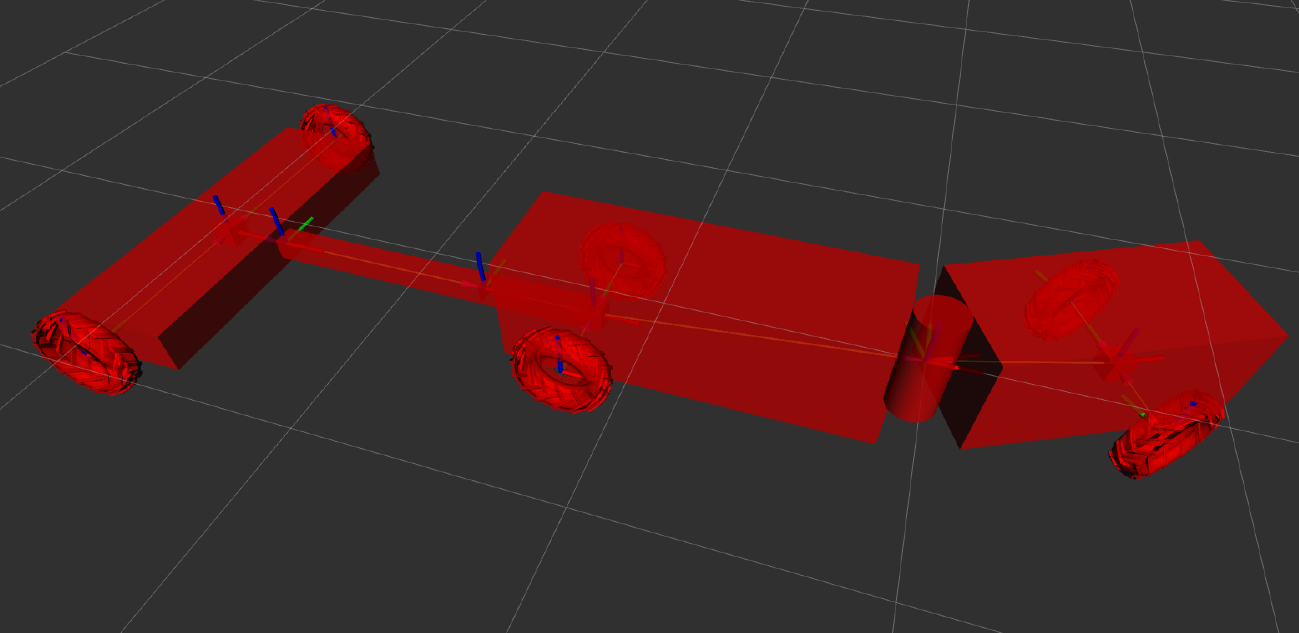}
    \label{fig:tractor-trailer-rviz}
    } \hspace{-1.em}
\subfigure{
    \includegraphics[width=0.47\columnwidth]{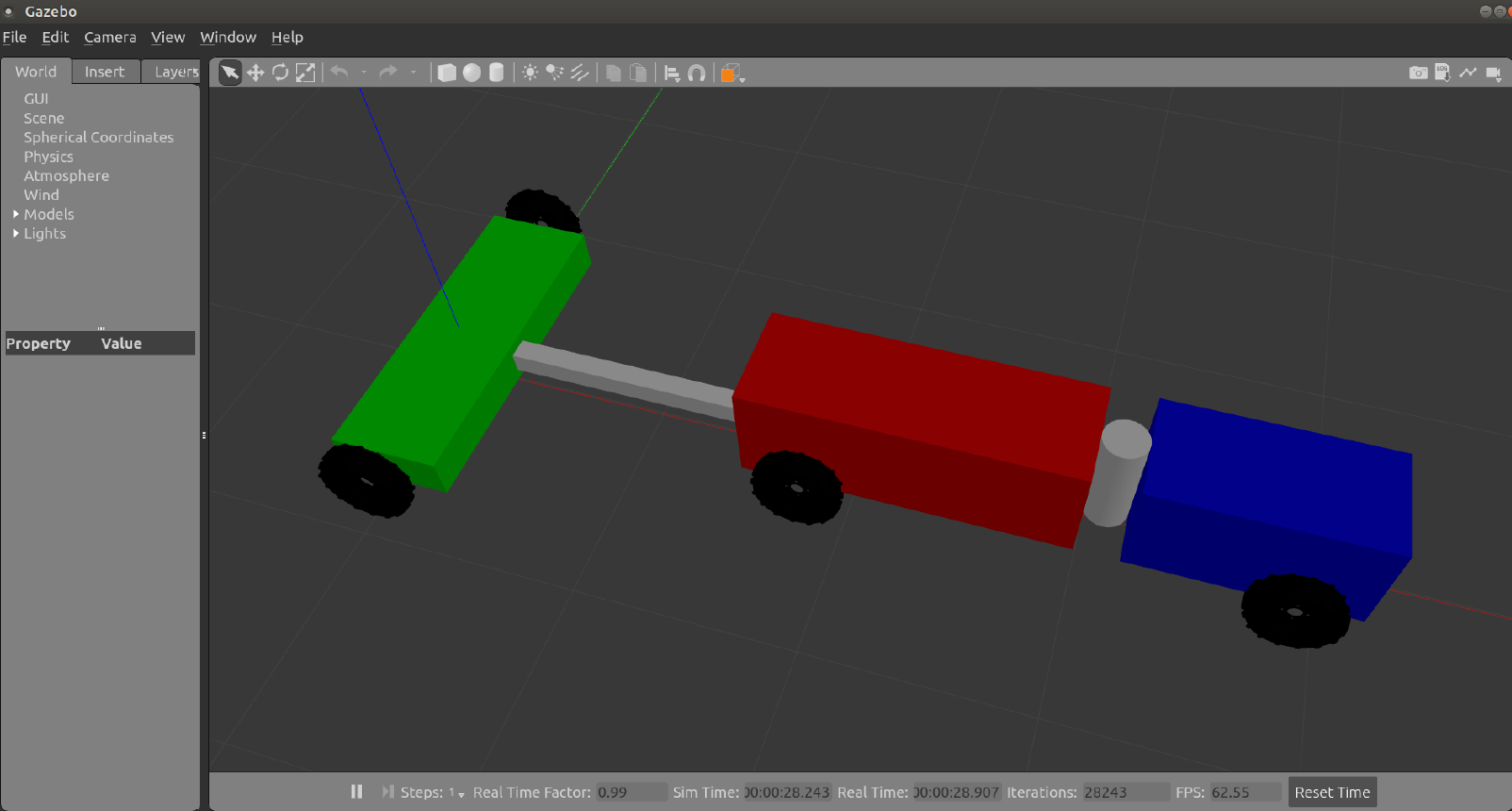}
    \label{fig:tractor-trailer-gazebo}
    }
\caption{Articulated tractor-trailer in RViz (left) and Gazebo simulator (right).}
\label{fig:tractor-trailer}
\end{figure}

To describe the articulated tractor-trailer system in a machine-readable way, we took advantage of the Robot Operating System (ROS\footnote{\url{http://www.ros.org/}}) as it provides a set of tools for describing and modeling our system in a very realistic way. The format for describing our articulated tractor-trailer system in ROS is the Unified Robot Description Format (URDF), which consists of an XML document in which we include not only the physical properties of our vehicle but also locations of sensors, visual appearance, links, transmissions, collisions of each part of the system and frictional characteristics of tyres. Another advantage of describing our model in this way is that our articulated tractor-trailer system can be easily integrated with Gazebo simulator (See Fig.~\ref{fig:tractor-trailer}). 

To simulate the vehicle within Gazebo, we must specify its joints. In order to control the speed, we need to define four velocity joints for the vehicle’s wheels. The attitude of the articulated tractor-trailer system is controlled through two position joints which command the front steering angle and the central articulation angle. In this way, for instance, the central articulation joint can be defined as shown in Definition 1, where we indicate that this joint should rotate (type revolute) along the \emph{z}-axis and we set its max-min bounds using the upper and lower limits tags.
\begin{Verbatim}[fontsize=\scriptsize,commandchars=\\\{\},formatcom=\color{black},frame=single,framesep=2mm,label=Definition 1: Central articulation joint,labelposition=bottomline]
<joint name="base_link__front_cradle_joint" type="revolute">
    <axis xyz="0 0 1" />
    <origin xyz="0 0 0" rpy="0 0 0" />
    <parent link="base_link" />
    <child link="front_cradle" />
    <limit effort="100.0" lower="-${M_PI/4}" upper="${M_PI/4}" velocity="1.0" />
</joint>
\end{Verbatim}
For every non-fixed joint, we need to specify a transmission, which tells Gazebo what to do with that joint. For example, to describe the relationship between the actuator and the central articulation joint, we need to set the transmission element as described in Definition 2, where we specify the transmission type and the joint where it is connected to.
\begin{Verbatim}[fontsize=\scriptsize,commandchars=\\\{\},formatcom=\color{black},frame=single,framesep=2mm,label=Definition 2: Central articulation transmission element,labelposition=bottomline]
<transmission name="base_link__front_cradle__transmission" type="SimpleTransmission">
    <type>transmission_interface/SimpleTransmission</type>
    <actuator name="base_link__front_cradle__motor">
        <hardwareInterface>hardware_interface/PositionJointInterface</hardwareInterface>
        <mechanicalReduction>1</mechanicalReduction>
        <motorTorqueConstant>10000</motorTorqueConstant>
    </actuator>
    <joint name="base_link__front_cradle_joint">
        <hardwareInterface>hardware_interface/PositionJointInterface</hardwareInterface>
    </joint>
</transmission>
\end{Verbatim}
To command the position of the central articulation joint, we need to set the hardware interface tag as a position joint interface in order to model the actuator as a servomotor. In a similar way, the position joint which commands the front steering can also be defined.

In order to describe wheels’ spinning, velocity joints are defined of continuous type, rotating along the \emph{y}-axis without any restrictions. For example, for the front left wheel, the joint should be defined as shown in Definition 3.
\begin{Verbatim}[fontsize=\scriptsize,commandchars=\\\{\},formatcom=\color{black},frame=single,framesep=2mm,label=Definition 3: Front left wheel joint,labelposition=bottomline]
<joint name="front_left_wheel" type="continuous">
    <parent link="front_left_ackermann_steering_link"/>
    <child link="front_left_wheel_link"/>
    <origin xyz="0 0 0" rpy="0 0 0" />
    <axis xyz="0 1 0" rpy="0 0 0" />
</joint>
\end{Verbatim}
To describe the relationship between the actuator and the velocity joint of the front left wheel, we set the transmission element as shown in Definition 4. 
\begin{Verbatim}[fontsize=\scriptsize,commandchars=\\\{\},formatcom=\color{black},frame=single,framesep=2mm,label=Definition 4: Front left wheel transmission element,labelposition=bottomline]
<transmission name="front_left_wheel_trans" type="SimpleTransmission">
    <type>transmission_interface/SimpleTransmission</type>
    <actuator name="front_left_wheel_motor">
        <hardwareInterface>hardware_interface/VelocityJointInterface</hardwareInterface>
        <mechanicalReduction>1</mechanicalReduction>
    </actuator>
    <joint name="front_left_wheel">
        <hardwareInterface>hardware_interface/VelocityJointInterface</hardwareInterface>
    </joint>
</transmission>
\end{Verbatim}
In this case, to model the actuator as a motor, we need to specify the hardware interface tag as a velocity joint interface so as to command its velocity, and hence, the speed of the vehicle.

It is worth mentioning that mass, inertia and wheel's friction properties are also considered in the model simulated by Gazebo. Our code is open source and it can be downloaded from our repository\footnote{\url{https://github.com/marinahmurillo/articulated_tractor_trailer_paper.git}}. We need to emphasize that we do not know how Gazebo simulates the behavior of the system at hand. However, we do know that in order to simulate the system dynamics, it accesses multiple high-performance physics engines such as ODE, Bullet, Simbody, and DART. As such, both the model simulated by Gazebo and the proposed mathematical model for the articulated tractor-trailer system are different. The latter is simpler, but for us is the best model at hand and, as it will be shown in the simulation example, even though it does not include any dynamic characteristics of the system, when it is used within the NMPC controller, it is enough to accurately control the trailer's position along the pre-defined path. It would be more accurate to include the dynamic characteristics of the articulated tractor-trailer system in the mathematical model. Nonetheless, this model would be somehow more difficult to obtain, it may result in larger state and control input vectors, leading to a higher computational cost; and, probably, simulation results would be similar to the ones we have obtained with a simpler model.

Parameters of the articulated tractor-trailer system are set accordingly as $L_f=0.8\,[\text{m}]$, $L_r=1.3 \,[\text{m}]$, $ d_1 = 0.5 \,[\text{m}]$ and $d_2 = 1.3 \, [\text{m}]$. Weight matrices are chosen as $Q = P = \operatorname{diag}([150, 300, 1, 100, 1, 100])$ and $R = \operatorname{diag}([25, 1, 1])$. The horizon and sampling period are set as $N=6 \, [\text{s}]$ and $T_s = 0.1 \, [\text{s}]$, respectively. In order to ensure that the resulting behavior of the system does not exceed the limitations of its actuators and mechanics, the following constraints are imposed: $|\gamma| \leq 60 \, [\text{deg}]$, $|\phi| \leq 60 \, [\text{deg}]$, $|v_f| \leq 2 \, \left[\text{m}/\text{s}\right]$, $|\Delta v_f| \leq 0.5  \, \left[\text{m}/\text{s}\right]$, $|\omega_1| \leq 15  \, \left[\text{deg}/\text{s}\right]$, $|\Delta \omega_1| \leq 10 \, \left[\text{deg}/\text{s}\right]$, $|\omega_2| \leq 15 \, \left[\text{deg}/\text{s}\right]$ and $|\Delta \omega_2| \leq 10 \, \left[\text{deg}/\text{s}\right]$. Continuous articulated tractor-trailer system defined in Eq.~(\ref{eq:sist_full_cin}) is discretized using collocation method with $3$ collocation points. In the following subsections, two simulation examples are shown. In the first scenario, the controller does not know that the trailer is towed to the articulated tractor-trailer system and, instead of controlling the position of the trailer itself, we control the \emph{xy}-position of the front block of the tractor, i.e. $x_f$ and $y_f$. In the second scenario, the controller is aware that the trailer is towed to the articulated tractor-trailer system and, hence, the goal is to control its position rather than the tractor’s. It should be pointed out that the objective function used in both examples is the same, the only difference is the mathematical model embedded in the NMPC controller.

With the goal of illustrating a possible outcome of a common practice in agriculture, the problem of using an articulated tractor-trailer system to seed a small $1600 \,[\text{m}^2]$ field is considered. It should be mentioned that, with the proposed vehicle model and the NMPC controller the articulated tractor-trailer system could follow almost any trajectory. The only limitation would be the feasibility of the path to be followed, i.e. it should take into account the physical limitations of the articulated tractor-trailer system.

\subsection{First example: controlling tractor's front block position}
In this first scenario, the controller is assumed to have no knowledge of the trailer kinematics, therefore, the front block of the tractor is required to follow the reference trajectory while expecting the trailer to travel approximately the same path.
\begin{figure}[h!]
\centering
    \includegraphics[width=0.9\columnwidth]{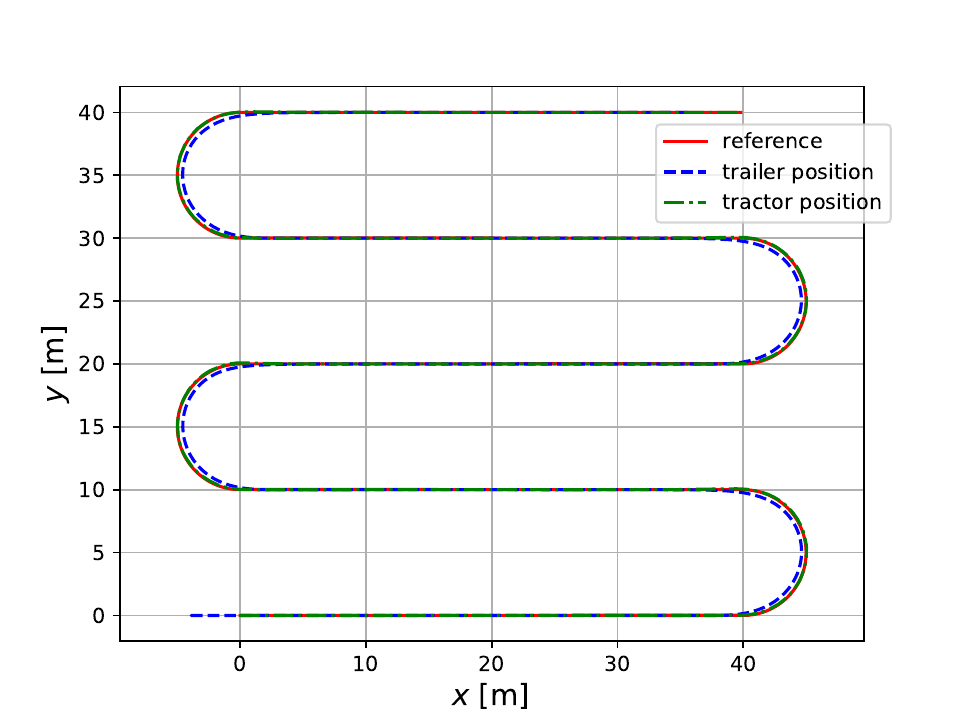}
\caption{Path traveled by the articulated tractor-trailer system when the tractor's front block position is controlled}
\label{fig:trajectory_tracking_tractor}
\end{figure}
As it can be seen in Fig.~\ref{fig:trajectory_tracking_tractor}, both the trailer and the tractor's front block follow the desired path accurately along straight paths. However, in headland turns only the tractor's front block follows the path accurately and the trailer describes a circumference of a smaller radii than the one described by the reference path. 
\begin{figure}[h!]
\centering
\subfigure{
    \includegraphics[width=0.47\columnwidth]{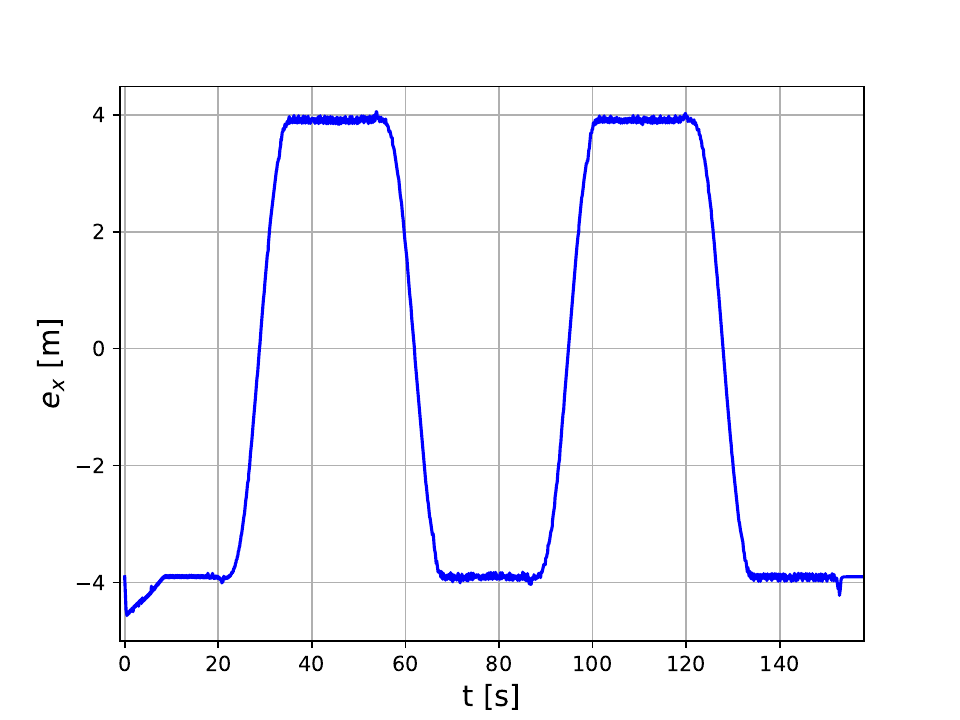}
    \label{fig:ex_tractor}
    }
\subfigure{
    \includegraphics[width=0.47\columnwidth]{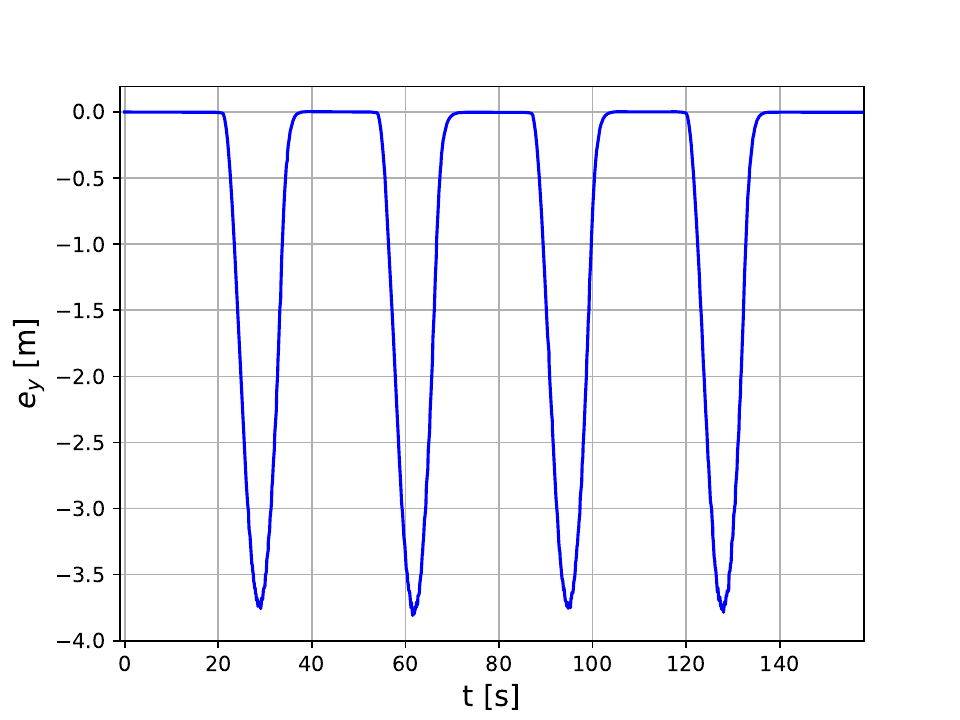}
    \label{fig:ey_tractor}
    }
\caption{Error deviation between reference path and tractor's front block position}
\label{fig:errors_tractor}
\end{figure}
In Fig.~\ref{fig:errors_tractor} errors $e_{x_{k|k}} = \mathbf{x}_{{\{x_t\}}_{k|k}} - \mathbf{r}_{\mathbf{x}_{{\{x_t\}}_{k|k}}}$ along \emph{x}-axis (left) and $e_{y_{k|k}} = \mathbf{x}_{{\{y_t\}}_{k|k}} - \mathbf{r}_{\mathbf{x}_{{\{y_t\}}_{k|k}}}$ along \emph{y}-axis (right) are depicted. It should be noted that when the vehicle moves alongside infield rows, $e_{x_{k|k}}$ it indicates that the trailer \emph{xy}-position is ahead or behind the desired path and it is related to acceleration and deceleration of the vehicle. On the other hand, it is essential to guarantee that the \emph{y}-position of the trailer remains as close as possible to the setpoint trajectory. Analyzing $e_{y_{k|k}}$, it can be seen that this error is very small when following straight paths while in headland turns this error is lesser than $3.8 \, [\text{m}]$. It is worth noting that, for instance, in a seeding process seeds and crops are planted alongside straight paths while in headland turns the implement, generally, is lifted up and no seeding occur in this part of the trajectory. To that end, more than reducing errors alongside the turning path, it should be more convenient to align the trailer both in the departure and the entrance of the infield paths. In this simulation example, the trailer is correctly aligned with the straight paths both at the end and the beginning of each infield row. However, as headland areas are generally restricted by physical dimensions it would be expected that the trailer position does not deviate too much from the desired trajectory.

\subsection{Second example: controlling trailer's position}
\begin{figure}[h!]
\centering
    \includegraphics[width=0.9\columnwidth]{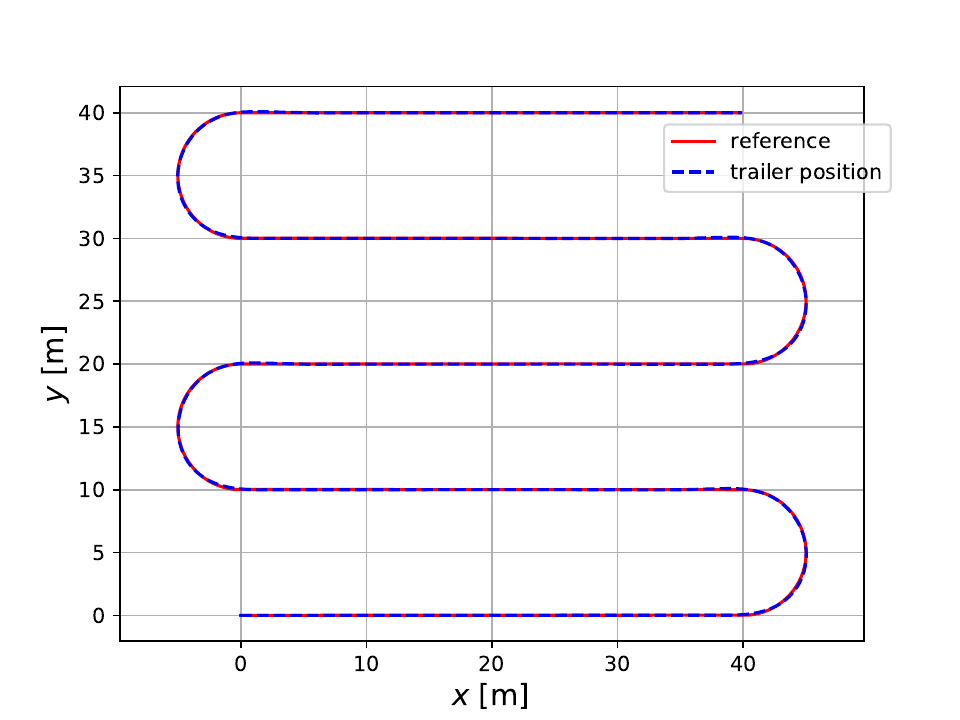}
\caption{Path traveled by the tractor-trailer system when the trailer is controlled}
\label{fig:trajectory_tracking}
\end{figure}

\begin{figure}[h!]
\centering
\subfigure{
    \includegraphics[width=0.47\columnwidth]{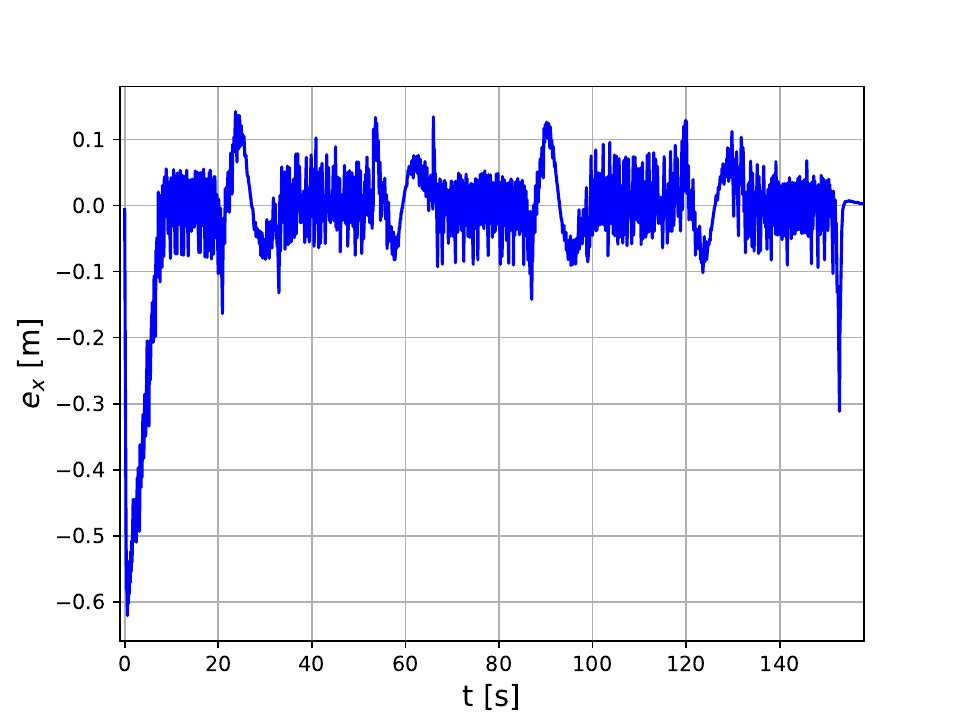}
    \label{fig:ex}
    }
\subfigure{
    \includegraphics[width=0.47\columnwidth]{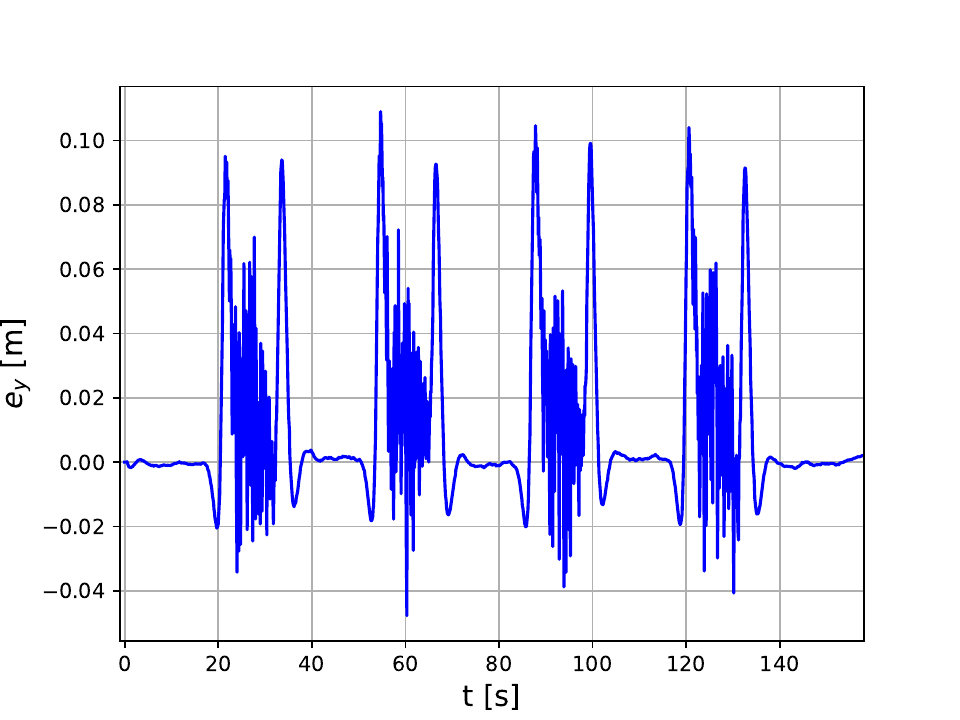}
    \label{fig:ey}
    }
\caption{Error deviation between reference path and trailer position}
\label{fig:errors}
\end{figure}

\begin{figure}[h!]
\centering
\subfigure{
    \includegraphics[width=0.48\columnwidth]{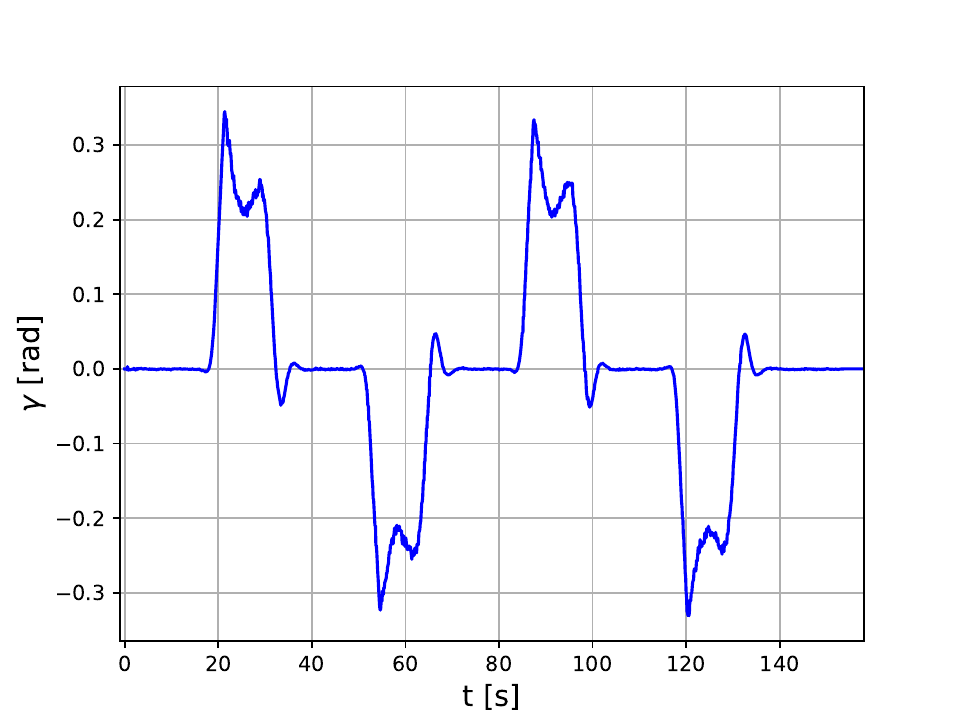}
    \label{fig:gamma}\hspace{-1.em}
    }
\subfigure{
    \includegraphics[width=0.48\columnwidth]{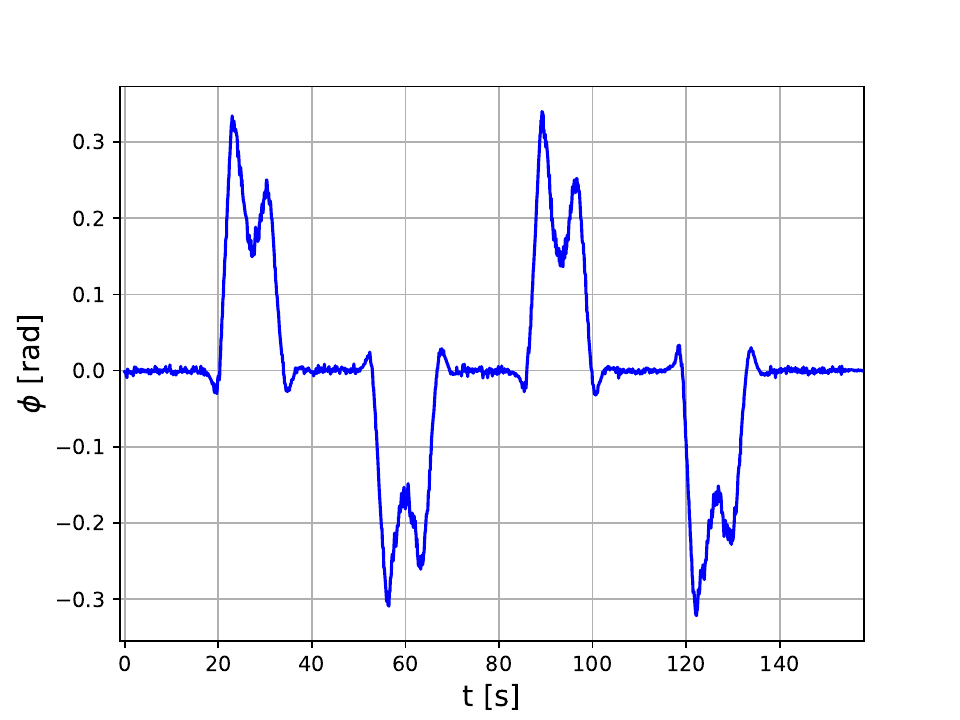}
    \label{fig:phi}
    }    
\caption{Articulation angle $\gamma$ (left) and steering angle $\phi$ (right)}
\label{fig:states_gamma_phi}
\end{figure}

In order to overcome the drawback of having large deviations alongside headland turns, we propose to perform the same simulation example as before but, this time, with our proposed articulated tractor-trailer system model. One of the main benefits of using this model is that the kinematics of the trailer can be embedded within the controller in an easy way, for instance, so that the trailer itself is able to follow the reference path. As it is shown in Fig.~\ref{fig:trajectory_tracking}, the trailer follows the desired path with a great accuracy not only along straight paths but also in headland turns. 
Figure~\ref{fig:errors} shows errors $e_{x_{k|k}}$ and $e_{y_{k|k}}$. The first one shows that $e_{x_{k|k}}$ is bigger at the beginning of the simulation but it decreases as the vehicle starts moving, leading to an error that is lesser than $16\, [\text{cm}]$ when following the desired trajectory. 
According to $e_{y_{k|k}}$, it can be seen that this error remains below $1 \, [\text{cm}]$ when following straight paths while in headland turns this error is lesser than $12 \, [\text{cm}]$, which is, for instance, much lower than that obtained in Fig.~\ref{fig:ey_tractor}. 
As it can be observed, by using an NMPC-based controller with our proposed articulated tractor-trailer system model, the vehicle is able not only to follow accurately straight paths until the end of each row but also it is able to enter the next row almost with no deviations. Furthermore, errors alongside turning paths can be substantially reduced if the trailer kinematics is taken into account in the NMPC-based controller.
Figure~\ref{fig:states_gamma_phi} depicts the evolution of articulation and steering angles, respectively. There, it can be seen that when the articulated tractor-trailer system moves within straight paths, both angles $\gamma$ and $\phi$ are approximately zero, thus allowing the vehicle to move forward without minor deviations along the \emph{y}-axis. When the vehicle reaches the end of a row, these angles start moving in a jointly way to successfully perform headland turns.

\begin{figure}[h!]
\centering
\subfigure{
    \includegraphics[width=0.3\columnwidth]{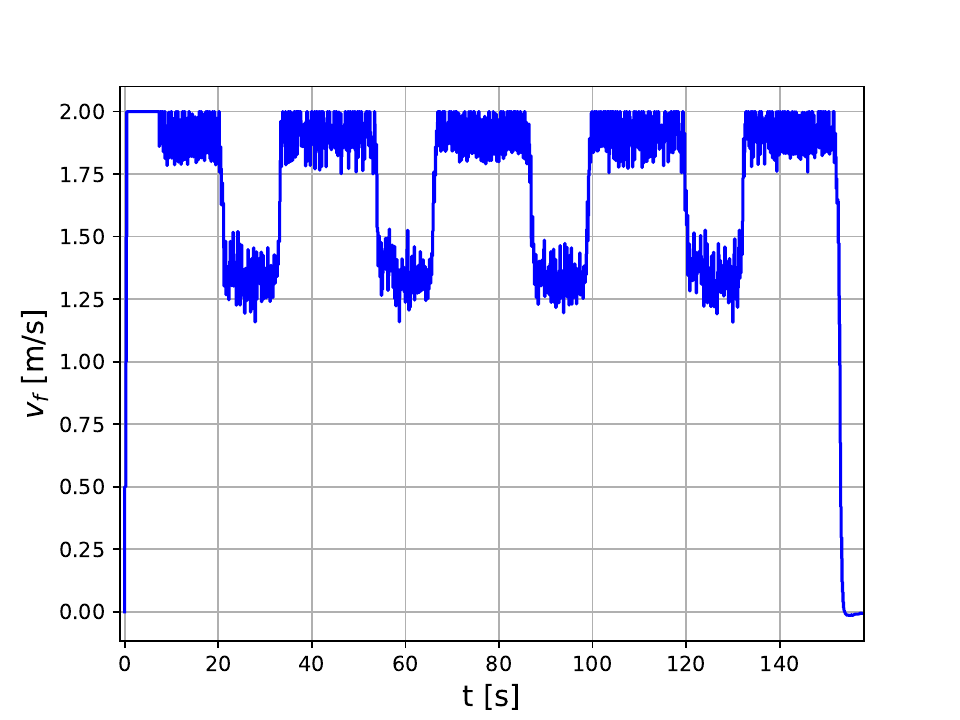}
    \label{fig:vf}\hspace{-1.em}
    }
\subfigure{
    \includegraphics[width=0.3\columnwidth]{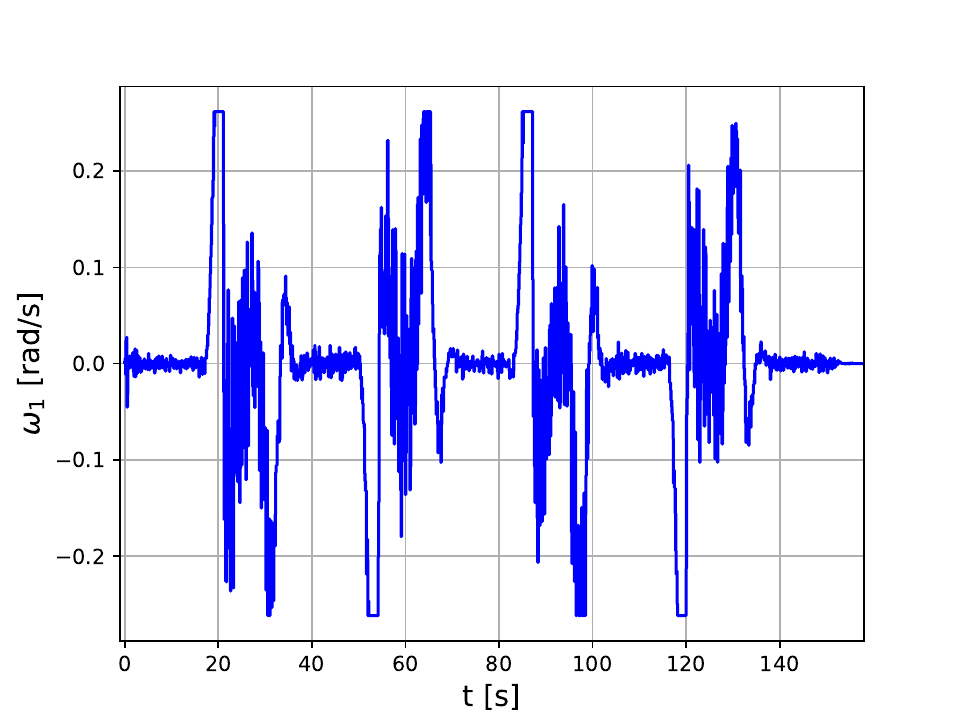}
    \label{fig:gamma_dot}\hspace{-1.em}
    }    
\subfigure{
    \includegraphics[width=0.3\columnwidth]{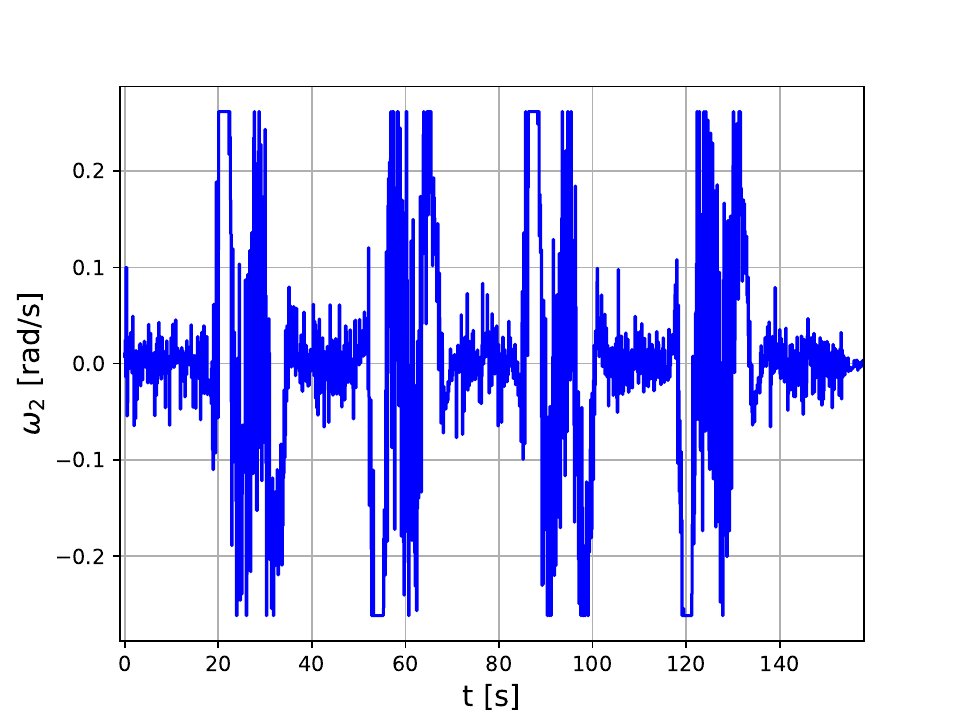}
    \label{fig:phi_dot}
    }    
\caption{Speed of the front block $v_f$ (left), angular velocity of articulation angle $\omega_1$ (middle), angular velocity of steering angle $\omega_2$ (right)}
\label{fig:control_inputs}
\end{figure}

Resulting control inputs are depicted in Fig.~\ref{fig:control_inputs}. As it can be seen, the velocity of the vehicle goes from zero to $2 \, \left[\text{m}/\text{s}\right]$, which is, for instance, the maximum bound we had set to this control input. When the vehicle is moving alongside straight paths, its speed oscillates between $1.75 \, [\text{m}/\text{s}]$ and the maximum speed, hence allowing to control the trailer's position more precisely. When the articulated tractor-trailer system is about to departure away from the infield row, its velocity is slowed down between  $1.15 \, [\text{m}/\text{s}]$ and $1.55 \, [\text{m}/\text{s}]$ in order to perform headland turns as close as possible to the reference trajectory. Angular velocities $\omega_1$ and $\omega_2$ are related to $\gamma$ and $\phi$, respectively, by time derivatives, and, as it can be observed in Fig.~\ref{fig:control_inputs} their time evolution is consistent with that obtained in Fig.~\ref{fig:states_gamma_phi}. The violent vibration that exhibit control inputs (Fig.~\ref{fig:control_inputs}) might not be realizable within practical implementations. To tackle this problem, one possibility would be to use the speed $v_f$ as a state variable (rather than a control input) and to describe it by a first or second order differential equation. In this way, the speed would show a smoother behavior than that shown in Fig.~\ref{fig:control_inputs} (left). On the other hand, the violent oscillation in both angular velocities $\omega_1$ and $\omega_2$ can be reduced in a similar manner. As it can be seen in the last two rows of Eq.~(\ref{eq:sist_full_cin}), the state equations for both $\gamma$ and $\phi$ are directly the associated angular velocities. Thus, in order to avoid high frequency oscillations, it would be possible to change these pure integrators by a first order differential equation of the form
\begin{equation}
    \dot{\gamma} = -k_1 \gamma + k_2 u_{\gamma} \hspace{0.3cm} \text{and} \hspace{0.3cm} \dot{\phi} = -k_3 \gamma + k_4 u_{\phi} 
\end{equation}
where $k_i$ (with $i=1,2,3,4$) denotes appropriate constants, $u_{\gamma}$ and $u_{\phi}$ are the control inputs associated to the states $\gamma$ and $\phi$, respectively.

\section{Discussion}
\label{sec:discussion}

The main goal of this research was to develop and to test the performance of a mathematical model of an articulated tractor-trailer system, which would be extremely suitable for PA purposes. For instance, it allows the the accurate path-tracking not only of the trailer's position but also of the tractor's one. Moreover, when the latter is monitored, although the trailer does not follow accurately the path alongside headland turns, it is indeed correctly aligned both in the departure and entrance of each infield row, decreasing errors within straight paths. Despite the fact that several works tackle the problem of controlling the tractor's and trailer's \emph{xy}-position \citep{pickett2016system,merx2017arrangement}, they mainly use independent controllers for both the tractor and the trailer, which might lead to deviation errors as the interaction between the tractor and the trailer might not be considered. To that end, we proposed to use a centralized approach in order to include this interaction in the design stage.

On the other hand, advanced control techniques such as NMPC have also been used to control tractor-trailer systems \citep{backman2012navigation, kayacan2014learning}. Nevertheless, vehicles reported in these works are restricted only to front steering and they do not include a central articulation joint. In this sense, our proposed mathematical model can be regarded as a generalization of those models with more limited steering mechanisms.

Even though areas covered by headlands turns are, in general, not used for seeding or harvesting issues, they are an essential part of the path-planning process as they comprise different restrictions such as time minimization, fuel efficiency and avoidance of restricted areas, among others, that should be included within the path-planning stage. Due to the fact that headlands areas are considered of low productivity, it is extremely important to minimize deviations alongside these turns. In our article, we do not tackle the problem of optimizing headland turns, however, we do consider its feasibility with respect to the physical capabilities of the articulated tractor-trailer system. Indeed, using our articulated tractor-trailer system model embedded within the NMPC controller, the \emph{xy}-position of the trailer can be monitored precisely and it can be maintained very close to the desired path, hence minimizing errors not only within straight paths but also along headland turns. It should be pointed out that we did not have to include extra information about turns, we only set the desired path and the controller itself adjusted control inputs in order to keep the trailer as close as possible to the desired path.

\section{Conclusion and future work}
\label{sec:concl}
In this work, an articulated tractor-trailer system with front steering has been studied. We showed that, by using a NMPC-based controller, Gazebo simulator and a ROS compatible architecture, the trailer managed to follow the desired path accurately. Indeed, the main advantage of using our proposed articulated tractor-trailer model is that the trailer's kinematics can be embedded within the NMPC controller, thus controlling the trailer's \emph{xy}-position is straightforward. Furthermore, it allows for precise trailer's path following not only alongside straight paths but also in headland turns. Despite the fact that, generally, the implement is lifted up when performing headland turns, it is extremely important to reduce the error in this area as they are mostly restricted by physical dimensions. On the other hand, our model allows for precise alignment of the trailer both in the departure and the entrance of the infield path, regardless the trailer kinematics is taken into account in the model itself or not. The future work of this research is aligned with the acquisition of a more precise mathematical model that considers the effect of non-flat terrains on the behavior of the system. The resulting model would exhibit a greater complexity, given that the angles of pitch and roll of each block of the vehicle would need to be taken into consideration and, hence, the controller would be able to compensate for their associated errors.

\section{Acknowledgments}
The authors wish to thank the \emph{Universidad Nacional de Litoral} (with CATT 2019 N° 17/01/2019), the \emph{Agencia Nacional de Promoci\'on Cient\'ifica y Tecnol\'ogica} (with PICT-2017-0543 and PICT-2016-0651) and the \emph{Consejo Nacional de Investigaciones Cient\'ificas y T\'ecnicas} (CONICET) from Argentina, for their support. A proper recognition should also be made to the teams that created CasADi and MPCTools, who have released them as open source software.

\bibliography{bibliography}

\end{document}